\documentclass[letterpaper, 10 pt, conference]{ieeeconf}  %
\usepackage{soul}
\usepackage{graphicx}
\usepackage{booktabs}
\usepackage{wrapfig}
\let\labelindent\relax
\usepackage{enumitem}
\usepackage{bm}
\usepackage{amsmath}
\usepackage{amssymb}
\usepackage{bbm}
\usepackage{color}
\usepackage{url}
\usepackage{dsfont}
\newcommand{\ours}{GET-Zero}

\definecolor{lightblue}{rgb}{.90,.95,1}
\sethlcolor{lightblue}

\IEEEoverridecommandlockouts                              %

\makeatletter
\let\NAT@parse\undefined
\makeatother
\usepackage{hyperref}  %

\title{\LARGE \bf
GET-Zero: Graph Embodiment Transformer \\ for Zero-shot Embodiment Generalization
}

\author{Austin Patel$^{1}$ and Shuran Song$^{1}$%
\thanks{$^{1}$Stanford University}%
\thanks{Project Page: \url{https://get-zero-paper.github.io}}%
}

\begin{document}

\maketitle
\thispagestyle{empty}
\pagestyle{empty}

\begin{abstract}

This paper introduces GET-Zero, a model architecture and training procedure for learning an embodiment-aware control policy that can immediately adapt to new hardware changes without retraining. To do so, we present Graph Embodiment Transformer (GET), a transformer model that leverages the embodiment graph connectivity as a learned structural bias in the attention mechanism. We use behavior cloning to distill demonstration data from embodiment-specific expert policies into an embodiment-aware GET model that conditions on the hardware configuration of the robot to make control decisions. We conduct a case study on a dexterous in-hand object rotation task using different configurations of a four-fingered robot hand with joints removed and with link length extensions. Using the GET model along with a self-modeling loss enables GET-Zero to zero-shot generalize to unseen variation in graph structure and link length, yielding a 20\% improvement over baseline methods. All code and qualitative video results are on our \href{https://get-zero-paper.github.io}{project website}.

\end{abstract}

\section{Introduction}
\label{sec:introduction}

Learning algorithms have significantly improved robots' ability to sense and adapt to external environments, yet most robots today cannot tolerate minor changes in their internal hardware. From missing links caused by damage to added joints for design improvements, hardware modifications often require retraining existing policies with embodiment-specific data. %

Recent methods develop embodiment-aware policies that condition on the hardware configuration of a robot design to control a class of embodiments~\cite{octo,sharedmotorpolicies,mybodyisacage,metamorph,manyquadrupeds,onepolicytorunthemall}, which enables efficient data reuse across embodiments. 
However, these method often have incomplete embodiment representations which ignore the graph connectivity of the robot, and thus cannot generalize well to an embodiment with a varied graph structure. 
Improving the performance under graph variations would greatly expand the applicability of embodiment-aware methods in domains such as manipulation, where the number of fingers may vary or in locomotion, where the leg may have a varying number of joints.

\begin{figure}[t]
    \centering
    \includegraphics[width=1\linewidth]{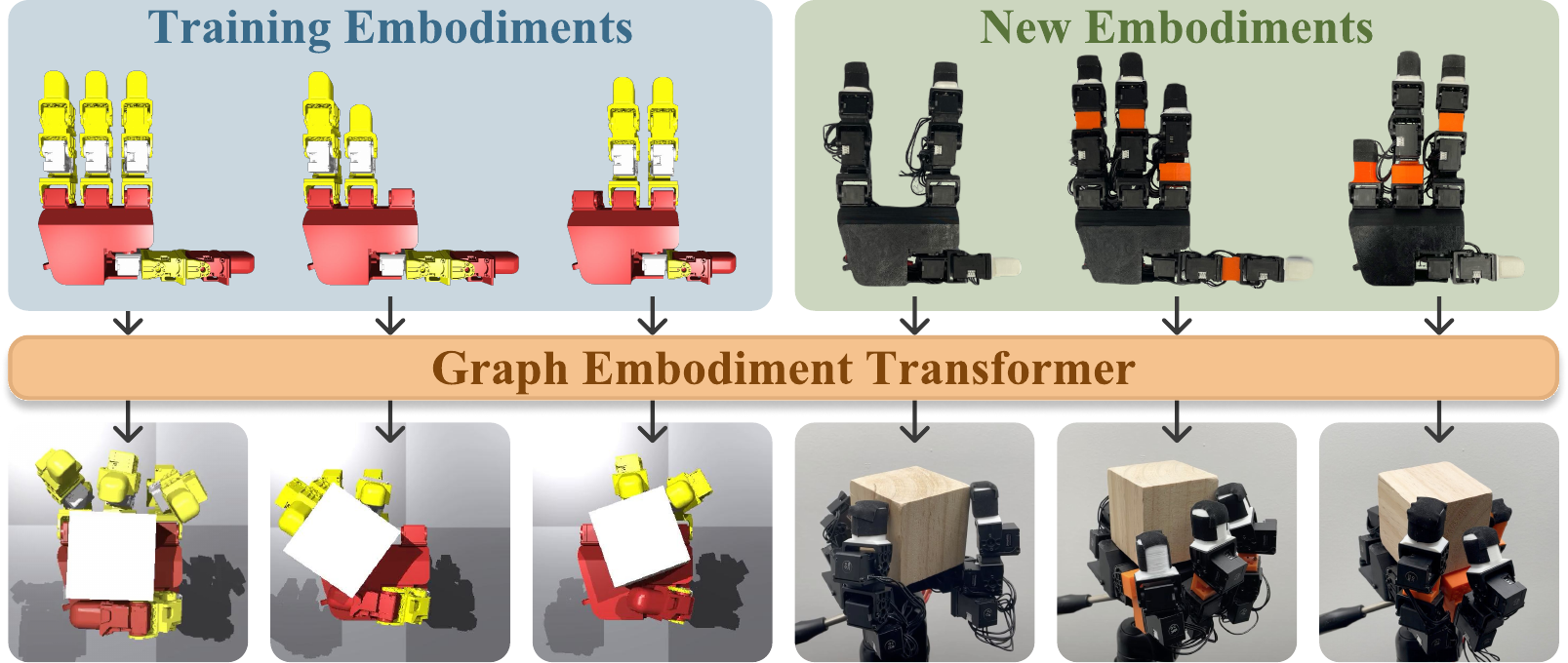}
    \caption{\textbf{GET-Zero} is an embodiment-aware policy that is able to zero-shot generalize to unseen embodiment designs with varied geometry, number of joints, and graph structure.}
    \label{fig:teaser}
    \vspace{-5mm}
\end{figure}

Motivated by this limitation, the goal of this work is to introduce an embodiment representation that explicitly leverages the graph connectivity of the robot to improve zero-shot control of robots with variations in graph structure. We propose \ours, which introduces the \textbf{Graph Embodiment Transformer (GET)} model architecture that enables \textbf{zero-shot adaptation} to new embodiments (Fig.~\ref{fig:teaser}). The key idea of GET is to leverage the robot graph connectivity as a structural bias in the transformer attention mechanism. By training this model using behavior data from a variety of robot embodiments, we can then zero-shot control previously unseen embodiments with varied geometry, number of joints, and graph structure. 
Specifically, \ours~consist of three components: 

\begin{enumerate}[leftmargin=4mm]
    \item \textbf{Graph Embodiment Transformer (GET).} GET is a modified transformer architecture~\cite{attentionisallyouneed} that encodes joints as separate tokens and uses the embodiment connectivity as a learned bias in the transformer attention mechanism. Attention enables local joint information to communicate and the graph attention bias modulates this communication based on how the joints are physically connected. 
    This extends prior embodiment-aware transformer models which either do not encode the embodiment connectivity~\cite{mybodyisacage} or have incomplete graph representations~\cite{metamorph}.
    
    \item \textbf{Embodiment-aware Distillation.} To train \ours~we first collect demonstration data from embodiment-specific experts. By conditioning on embodiment information, we then distill knowledge from embodiment-specific experts into an embodiment-aware GET model using behavior cloning (BC) that controls both training and unseen embodiments with a \textit{single} set of network weights. 
    Prior methods \cite{sharedmotorpolicies,mybodyisacage,metamorph} use RL to simultaneously learn an embodiment-aware model while learning to complete the task across all embodiments. In contrast, our method simplifies data generation by independently learning experts, then jointly distilling expert behavior.

    \item \textbf{Self-modeling Loss.} During the distillation BC phase, we introduce a self-modeling loss to predict the position of each joint in 3D space (i.e., forward kinematics). We found that this simple and general supervision improves zero-shot performance. Self-supervision is common in NLP \cite{bert} and vision \cite{vit} domains, but is less explored in cross-embodiment learning \cite{unsupervisedpretraining,zakka2021xirl,xu2023xskill}.

\end{enumerate}

To validate \ours, we conduct a case study to learn an embodiment-aware, dexterous in-hand object rotation policy across different hardware configurations of a multi-fingered robot hand. In particular, we use the LEAP Hand~\cite{leaphand}, which is a low-cost, four-fingered hand with four joints per finger consisting of 3D printed components and off-the-shelf motors. We create 44 hardware configurations of this hand by removing different combinations of finger joints and associated links. Next, we train embodiment-specific expert policies for each hand using RL in simulation and then follow the \ours~training procedure to distill the expert behavior into an embodiment-aware GET model. Our experiments in simulation and real world demonstrate how the novel graph encoding and self-modeling in \ours~improves the capability of this model to zero-shot control hand designs with unseen variations in link geometry and graph connectivity.

\section{Related Work}
\label{sec:related_work}

\textbf{Cross-Embodiment pretraining with embodiment-specific finetuning.}
Recent approaches leverage large-scale pretraining, either on visual data, robot trajectories, language, or tasks as an initialization for policy learning or generalist agents~\cite{rt22023arxiv, radosavovic2023robot, ageneralistagent, ageneralistdynamicsmodel, moo2023arxiv, driess2023palme, kalashnikov2018scalable}. Other approaches leverage cross-embodiment robot data to initialize a generalist agent, and then fine-tune the base model or an action head to adapt to new robot embodiments~\cite{open_x_embodiment, bousmalis2023robocat}.
These method often assume a unified action space, which is reasonable for navigation \cite{shah2023vint} or for robot arm policies where the end-effector pose is a shared representation across arms \cite{bousmalis2023robocat, yang2024pushing,open_x_embodiment, variableimpedancecontrol}. However, unified actions spaces are often not sufficient for tasks where precise, low-level joint actions matter, such as dexterous manipulation \cite{dexterous1,dexterous2,dexterous3,dexterous4}. It is challenging to formulate a unified action space in the case when hands have varying number of fingers and kinematic feasibility and when contact may occur anywhere along the fingers. Instead, our method uses the robot graph structure as an embodiment representation to perform joint-level control for tasks where a unified action space is not possible. \ours also does not assume demonstration data on the target embodiment is available nor require finetuning.

\textbf{Embodiment Representation.}
Prior multi-embodiment policies demonstrate using a single policy to control different robot designs \cite{octo, unigrasp, sharedmotorpolicies, mybodyisacage, metamorph, onepolicytorunthemall, manyquadrupeds}. %
However, most methods show limited generalization ability to new embodiment designs beyond the training set due to incomplete embodiment representations. For example, Octo~\cite{octo} has 3rd-person images of the robot arm which doesn't generalize across robot appearance changes, though this could be addressed with robot inpainting~\cite{chen2024mirage}. Other methods learn cross-embodiment representations that align actions across different embodiments~\cite{xu2023xskill, chen2019hardwareconditionedpolicies, zakka2021xirl}. In contrast, \ours uses an embodiment graph composed of joints which is readily accessible from URDF files as a practical representation when joint-level actions matter.

\textbf{Embodiment-aware policy architectures.}
Most related to our approach is the line of work developing embodiment-aware policies. 
Prior methods~\cite{nervenet,learningtocontrolselfassembling,sharedmotorpolicies} structure a graph neural network (GNN) to match the embodiment graph with joints as nodes and links as edges to control stick-like walking characters in simulation. \cite{mybodyisacage} shows that a GNN matching the embodiment graph is outperformed with the fully connected attention mechanism in transformers~\cite{attentionisallyouneed} despite having no graph encoding. \cite{metamorph} extends this model with dynamics and kinematics encodings, demonstrating zero-shot generalization to these properties. However, these methods do not generalize well to unseen embodiment graphs due to no~\cite{mybodyisacage} or limited~\cite{metamorph} graph representations. Our method \ours~extends the transformer architecture in~\cite{mybodyisacage,metamorph} to include an explicit graph representation that improves zero-shot generalization to unseen embodiment graphs. Furthermore, unlike prior embodiment-aware RL methods that do locomotion only in simulation~\cite{nervenet,learningtocontrolselfassembling,sharedmotorpolicies,mybodyisacage,metamorph}, \ours~uses behavior cloning from experts for an in-hand rotation task transferable to a real robot.

\section{Method: Graph Embodiment Transformer}
\label{sec:model}

\begin{figure*}[t]
    \centering
    \includegraphics[width=0.75\linewidth]{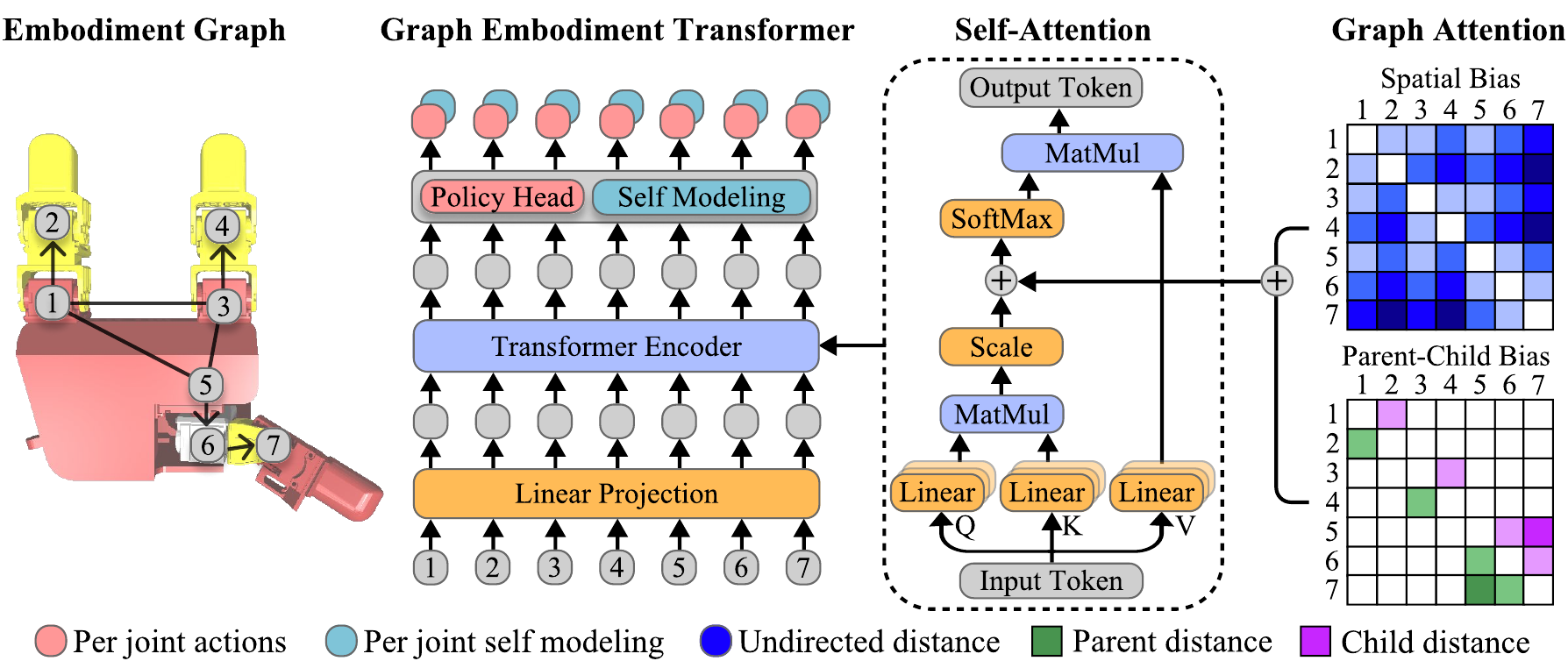}
    \caption{\textbf{Graph Embodiment Transformer (GET).} GET is an embodiment-aware model based on a transformer encoder. Each joint forms separate tokens containing local sensory and embodiment information. The self-attention layers use an undirected (Spatial Bias) and directed (Parent-Child Bias) graph distance to bias the attention scores between joints according to the embodiment graph (grid color intensity indicates distance between nodes). A policy head predicts actions and a self modeling head predicts meta-properties about the embodiment, such as forward kinematics.} \vspace{-3mm}
    \label{fig:architecture}
\end{figure*}

We present Graph Embodiment Transformer (GET), an embodiment-aware transformer encoder~\cite{attentionisallyouneed} that uses the embodiment graph as a structural bias in network architecture (Fig.~\ref{fig:architecture}). 
An embodiment graph consists of nodes representing local joint information, directed edges representing links connecting parent and child joints, and undirected edges between joints at the start of independent serial chains.
GET represents each joint as separate transformer tokens containing both local sensory observations and local hardware properties (\S\ref{sec:token_format}). The graph edges (i.e., links) are encoded through a learned attention bias in the self-attention layers (\S\ref{sec:graph_encoding}).

\subsection{Embodiment Tokenization}
\label{sec:token_format}
For a robot with $J$ joints, there are $J$ input tokens to the transformer encoder containing local observations and $J$ corresponding output tokens for per-joint actions. The encoder supports a variable number of tokens which enables compatibility as the number of joints varies across embodiments.

Observations can either be local $l$ to specific joints or global $g$ across the entire embodiment. Additionally, observations are either fixed $f$ or variable $v$ if they change during execution. This yields four observation types: variable local $o_{vl}$ (e.g., joint angle/velocity), variable global $o_{vg}$ (e.g., a global time encoding, or environment state), fixed local $o_{fl}$ (e.g., joint position in rest pose, or joint limit ranges) and fixed global $o_{fg}$ (e.g., a task identifier).
$H$ past history steps, indicated $t\rightarrow t-H$, are included for variable observations. The local observations for joint $j$ are defined as $o_{*l,j}$. The transformer token $T_j$ for joint $j$ at timestep $t$ is constructed as $T_{j}^{t} = [o_{vg}^{t\rightarrow t-H}, o_{fg}, o_{vl,j}^{t\rightarrow t-H}, o_{fl,j}]$. The tokens pass through a learned linear embedding before entering the transformer encoder.

\subsection{Graph Encoding}
\label{sec:graph_encoding}
One challenge is that the transformer encoder has no direct graph encoding mechanism in the original implementation~\cite{attentionisallyouneed}. Input tokens are permutation invariant without a positional encoding mechanism, but are often linearly encoded with sinusoidal or learned positional encodings. Prior embodiment-aware methods address this in multiple ways. \cite{mybodyisacage} use no positional encoding meaning no graph representation is present. \cite{metamorph} linearizes the graph using a depth-first search ordering then apply a learned linear positional encoding. However, this approach is sensitive to graph variations as the DFS ordering is not unique, which empirically caused a $\sim75\%$ drop in performance with the opposite node order in~\cite{metamorph}. \cite{bodytransformer} use the adjacency matrix as a binary attention mask to limit communication early in the transformer to adjacent nodes.

Our GET model uses a learned attention bias in the encoder self-attention mechanism to explicitly encode the graph similar to the Graphormer work by~\cite{graphormer}. We opt for Graphormer for its simple implementation, but note that our approach would benefit as graph transformers improve~\cite{graphgps,san}. Using the notation from~\cite{graphormer}, the inputs to the self-attention layer given $J$ joints are the hidden states $H=[h_1^\top,...,h_J^\top]\in\mathbb{R}^{J\times d}$. $H$ is projected by three matrices $W_Q\in\mathbb{R}^{d\times d_K}$,$W_K\in\mathbb{R}^{d\times d_K}$, $W_V\in\mathbb{R}^{d\times d_V}$ to produce $Q=HW_Q$, $K=HW_K$, $V=HW_V$. Then the attention layers will compute an attention matrix $A\in\mathbb{R}^{J\times J}$ where $A_{ij}$ represents the attention score computed between joints $i$ and $j$ as follows:
$A=QK^\top/\sqrt{d}, \text{Attn}(H)=\text{softmax}(A)V$.

To encode the embodiment graph, we learn two separate biases to the attention scores ($A$):

\textbf{Spatial Encoding.} The spatial encoding computes the shortest path distance (SPD) $\phi^\text{SPD}$ between node $i$ and node $j$ as $\phi^\text{SPD}(i,j)$, treating all edges as undirected. An embedding table $s$ is indexed with the SPD distance as $s_{\phi^\text{SPD}(i,j)}$ to get a learned scalar that is added to the attention score $A_{ij}$.

The intuition is that the attention mechanism can learn to pay more or less attention to joints in the graph based on their distance. For example, nearby joints may learn a large attention bias, while far away joints that may not have much influence on each other hence a low spatial bias.

\textbf{Parent-Child Encoding.} Unlike Graphormer \cite{graphormer} that only encodes undirected graph in the attention mechanism, we introduce a new parent-child attention bias to encode directed features. For joints $i$ and $j$, we compute the parent distance $\phi^P(i,j)=\phi^\text{SPD}(i,j)\mathds{1}\{i\text{ is parent of }j\}$ which is the distance between $i$ and $j$ if $j$ is the child of $i$ at some distance along the forward kinematic chain, otherwise $0$. We compute an analogous formula for the child distance as $\phi^C(i,j)=\phi^\text{SPD}(i,j)\mathds{1}\{i\text{ is child of }j\}$. There are associated scalar embedding tables $p$ and $c$. This means that the directed edges have distinct attention biases if we go from parent-to-child or child-to-parent.

Our motivation is that the parent-child encoding lets the model modulate attention based on directed connectivity of the forward joint chain, which could be helpful to compute forward kinematics.

\textbf{Attention Bias Computation.}
For each attention head and encoder layer, we add attention bias embeddings $s$, $p$, and $c$ to compute attention as: $A_{ij}=\frac{(h_iW_Q)(h_jW_K)^T}{\sqrt{d}} + s_{\phi^\text{SPD}(i,j)} + p_{\phi^\text{P}(i,j)} + c_{\phi^\text{C}(i,j)}$.

One key aspect of this approach is that this attention bias is invariant to the ordering of the input tokens unlike the depth-first linearization approach~\cite{metamorph}.

\textbf{Output Heads.}
After passing $J$ tokens through the transformer encoder, GET produces $J$ corresponding features. We include a policy head to predict actions and a self-modeling head used to predict meta-features about the embodiment. More details on the heads are discussed in \S\ref{sec:distill_experts}.

\section{Case Study: In-hand Object Rotation}
\label{sec:method}
\begin{figure}[t]
    \centering
    \vspace{1mm}
    \includegraphics[width=\linewidth]{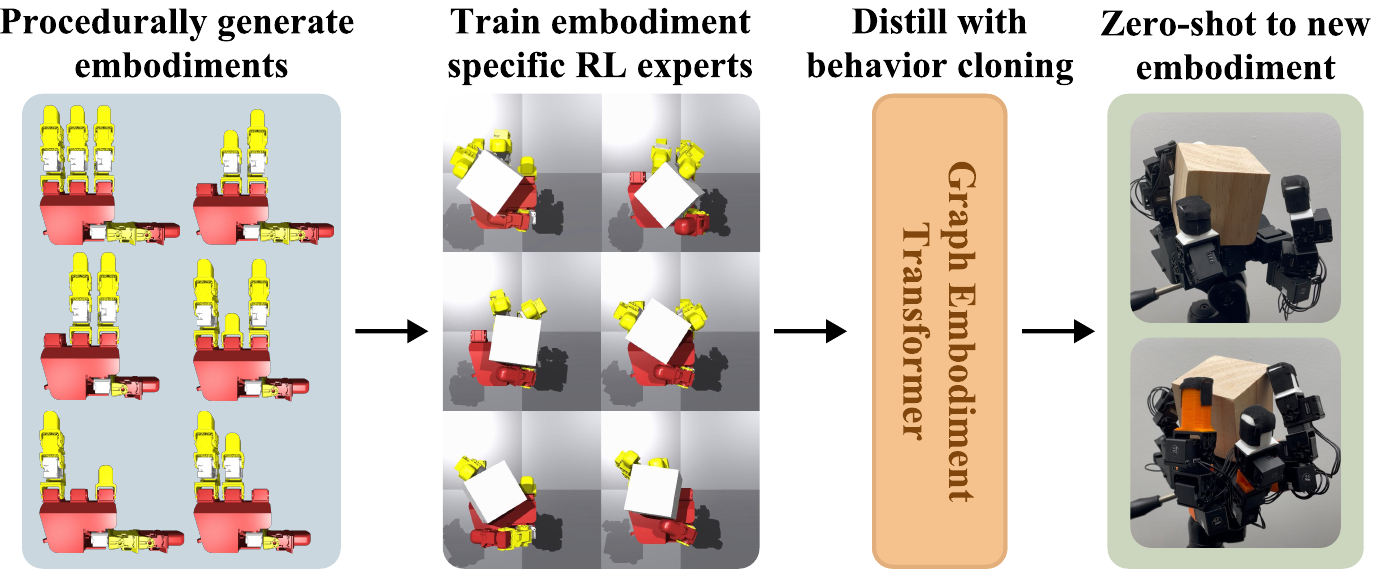}
    \vspace{-6mm}
    \caption{\textbf{Training procedure.} To train \ours, we follow a teacher-student paradigam, where the teachers are \textit{separate} embodiment-specific experts trained using RL. Then we distill knowledge from the experts into a \textit{single} embodiment-aware transformer (i.e., the student policy) using behavior cloning with a self-modeling loss. \ours~takes as input embodiment definition and proprioception, and infers proper actions to perform an in-hand rotation task for an unseen embodiment.}   
    \label{fig:method}
    \vspace{-4.5mm}
\end{figure}

We conduct a case study on applying \ours~to in-hand object rotation with different variations of LEAP Hand~\cite{leaphand}. 
Unlike pick-and-place tasks where a unified action space across different robot arms is easily defined (e.g., end-effector space), in-hand object rotation requires the policy to differentiate low-level joint space actions that are aware of the embodiment structure.  
Applying \ours~to in-hand rotation requires three stages (Fig.~\ref{fig:method}): 1) generate embodiment variations~(\S\ref{sec:procedural_generation}), 2) train RL embodiment-specific experts~(\S\ref{sec:rl_experts_training}), and 3) distill knowledge from experts into an embodiment-aware GET architecture~(\S\ref{sec:distill_experts}).

\subsection{Procedural Embodiment Generation}
\label{sec:procedural_generation}
\begin{figure}[t]
    \centering
    \vspace{1.5mm}
    \includegraphics[width=1\linewidth]{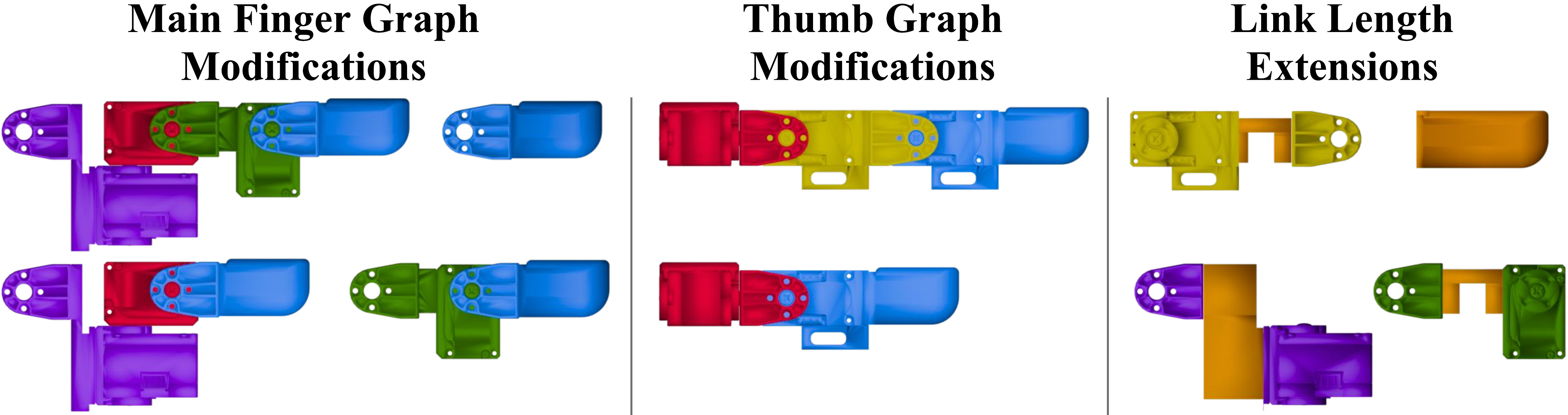}
    \caption{\textbf{Finger Variations.} We procedurally generate variations of the fingers in the LEAP Hand~\cite{leaphand} by removing joints and links as well as adding in 1.5cm link length extensions (orange).}
    \label{fig:finger_variations}
    \vspace{-5.5mm}
\end{figure}

The LEAP Hand~\cite{leaphand} consists of 3D printed components and off-the-shelf motors, making it highly configurable. The hand features three main fingers which are identical in structure and a thumb which has a separate structure. For the main fingers, we construct five variations of the fingers (4 combinations + no finger config) and two different variations of the thumb by removing various combinations of joints (Fig.~\ref{fig:finger_variations}). This yields $5^3\times2=250$ embodiment configurations from which we require 1) at least two fingers have at least one joint and 2) that there is at least one main finger with two joints, yielding 236 graph variation embodiments. We additionally introduce 1.5cm link length extensions shown in orange in Fig.~\ref{fig:finger_variations} to the 236 designs to generate additional hand designs.

\subsection{Train Embodiment-Specific Experts}
\label{sec:rl_experts_training}

We train embodiment-specific RL experts using PPO~\cite{ppo} in Isaac Gym~\cite{isaacgym} to complete the in-hand rotation task, adapted from the setup in \cite{leaphand}. 
This setup has rewards for rotation along the yaw axis, penalties for high motor torques and dropping the cube, and domain randomization across cube sizes. These learned policies operate only on proprioceptive joint states and predict delta joint targets that guide a PD controller. 
We pick the best of five RL seeds per embodiment and filter embodiments that do not complete a full $2\pi$ rotation within $30$s. Many hand designs can not complete the task under this threshold due to many missing joints, resulting in 44 training embodiments with associated expert policies. In general, it's not necessary to train experts for the entire design space, which might be large, but instead on a subset of representative embodiments (see Fig.~\ref{fig:training_embodiment}).

\subsection{Distill Experts into Embodiment-Aware Transformer}
\label{sec:distill_experts}

By rolling out embodiment-specific experts we can collect demonstration data for each hand. 
Specifically, we collect 7 hours worth of demonstrations for each embodiment. This demonstration data is combined with embodiment information to form the training dataset of (state, action, embodiment) tuples used to train the embodiment-aware GET model with behavior cloning (BC).

Separating collection of demonstration data from training embodiment-aware policies gives freedom to use existing policy architecture or training setups to generate expert demonstrations (e.g., teleoperation, RL, optimization methods, privileged information, etc.). Prior embodiment-aware methods~\cite{mybodyisacage,metamorph} instead jointly learn an embodiment-aware policy along with learning how to control training embodiments. This means the baselines do not support reusing existing demonstration data and require advanced replay buffer balancing as embodiments learn at different rates~\cite{metamorph}.

\textbf{GET Training Setup}. We train the GET architecture (\S\ref{sec:model}) using BC. Specifically, our token input has normalized joint angles and PD target joint states ($o_{vl}$), 3D joint position and joint rotation with respect to the parent joint from the URDF file ($o_{fl}$), and a sinusoidal phase encoding with a period of two seconds to encode cyclic progression of the rotation cycle ($o_{vg}$). The directed embodiment graph is an additional input used in the graph attention encoding (\S\ref{sec:graph_encoding}).

\textbf{Policy Head.} A policy head predicts per-joint single-step action. Each joint has a PD target state that is initialized to the starting joint state. The action head predicts delta actions that are added to this target state, which sets the target state for the low-level PD controller. We supervise these actions using $L_2$ loss with the demonstration data.

\textbf{Self-Model Head.} A self-modeling head predicts the current 3D pose of each joint in the robot local frame. This forward kinematics (FK) task is simple and generally applicable across embodiments, yet requires the network to use the spatial relationships from the graph and current joint angles along the FK chain. We supervise with the ground truth FK pose with $L_2$ loss. Both the policy and self-modeling heads operate on the same output latent representations to encourage actions to be predictable from an embodiment-rich representation. We opt for per-joint FK for its simplicity, but note that more complex self-modeling is possible \mbox{\cite{visualselfmodeling}}.

\section{Experiments}
\label{sec:experiments}

We procedurally generate 236 graph variations of the LEAP hand~\cite{leaphand} by removing various combinations of joints and associated links from the embodiment (\S\ref{sec:procedural_generation}). From these, we obtain 44 embodiments with graph variations with associated expert RL policies that achieve a baseline level of performance (\S\ref{sec:rl_experts_training}). These 44 embodiments serve as training embodiments and only have graph variations (link length extensions are not seen during training). Our experiments consist of using demonstration data from these embodiments to train various baselines and ablations of the GET architecture to evaluate zero-shot performance to unseen embodiment graphs and link extensions.

\textbf{Comparisons.} MetaMorph~\cite{metamorph} and Amorpheus~\cite{mybodyisacage} models are two prior embodiment-aware transformer methods. These methods also use a transformer but do not contain a full graph encoding or self modeling loss. We re-implement the positional encoding aspects of these method in our setup to focus on comparing the impact of our graph representation and self-modeling rather than differences in task (locomotion v.s. manipulation), training procedure (RL v.s. BC) or observation format. 

\begin{itemize}[leftmargin=3mm]
    \item \textbf{ET.} The Embodiment Transformer (ET) matches the architecture and token format of GET except without the graph encoding (\S\ref{sec:graph_encoding}) or the self modeling loss. This ET baseline most similarly matches the encoding method in Amorpheus~\cite{mybodyisacage}, which uses no graph encoding. This means that the model has no explicit connectivity information about the joints.

    \item \textbf{ET+DFS.} This baseline uses positional encoding from MetaMorph~\cite{metamorph} that linearizes the embodiment graph in depth-first search (DFS) ordering and applies a learned linear positional encoding.
\end{itemize}

We also ablate different components of our method such as self-modeling loss (SL), spatial encoding (SE), parent-child encoding (PE).  Results are summarized in Tab. \ref{tab:simulation}. 

\textbf{Evaluation Metric.} The task performance is measured by average rotational velocity along the yaw axis in degrees/second including standard deviation across five seeds.  We compute results by rolling out the embodiment-aware policies in a physics simulator for a total of 42 minutes of execution time per embodiment per seed. Each category averages over 10 unseen embodiments (see website for embodiment images).

Our expectation for zero-shot graph variation embodiments (given 44 training embodiments) is that joint sequences composing individual finger designs will have been seen before, but that 1) the finger position within the hand will be unseen (three positions corresponding to the three main fingers) and/or that 2) the combinations of finger designs across the four fingers will be unseen. We were able to train embodiment-specific RL experts for all 10 embodiments with graph variations with sufficient performance ($2\pi$ rotation within 30 seconds), validating that they are reasonable targets for zero-shot generalization.

\begin{table}[t]
    \centering
    \setlength{\tabcolsep}{0.5mm}
    \vspace{2mm}
    \caption{\textbf{Simulation results (average rotational velocity deg/sec).}  RL experts are embodiment-specific policies. ET: Embodiment transformer, DFS: depth-first graph encoding,\\SL: self-modeling loss, SE: spatial graph encoding,\\
    \vspace{-1mm}
    PE: parent-child graph encoding
    }
    \vspace{-2.5mm}
    \resizebox{\columnwidth}{!}{
    \begin{tabular}{l|cccc}
    
    \toprule
                              & Training Graph         & New Graph         & ~~New Geo~~           & New Graph\&Geo\\
    \toprule
     RL experts               & 16.50             & ---               & ---               & --- \\
     \toprule
ET \cite{mybodyisacage}                & 14.82$\pm$0.15                        & 8.68$\pm$0.34                         & 12.92$\pm$0.36                        & 8.12$\pm$0.46 \\                 
ET+DFS~\cite{metamorph} & 15.03$\pm$0.37                        & 6.45$\pm$0.28                         & 14.37$\pm$0.34                        & 6.27$\pm$0.22 \\                      
\midrule
ET+SL                                  & 14.68$\pm$0.33                        & 8.19$\pm$0.74                         & 12.71$\pm$0.72                        & 7.60$\pm$1.12 \\                      
                      
ET+PE+SE                                & 16.03$\pm$0.19                        & 9.65$\pm$0.23                         & 15.30$\pm$0.29                        & 9.05$\pm$0.42 \\
ET+PE+SL                         & 16.10$\pm$0.22                        & 9.56$\pm$0.47                         & 15.59$\pm$0.32                        & 8.81$\pm$0.42 \\                        
ET+SE+SL                               & 16.24$\pm$0.21                        & 10.04$\pm$0.15                        & 15.74$\pm$0.25                        & \textbf{9.75$\pm$0.20} \\                      
ET+PE+SE+SL                    & \textbf{16.32}$\bm{\pm}$\textbf{0.24} & \textbf{10.07}$\bm{\pm}$\textbf{0.58} & \textbf{15.80}$\bm{\pm}$\textbf{0.29} & \textbf{9.75}$\bm{\pm}$\textbf{0.54} \\
     \bottomrule
    \end{tabular}
    }
    
    \label{tab:simulation}
    \vspace{-6.5mm}
\end{table}

\begin{table*}[t]
    \normalsize
    \centering
    \setlength{\tabcolsep}{5pt}
    \vspace{1.61mm}
    \caption{\textbf{\ours~sim-to-real evaluation.} [Col 1] An AR tag tracks cube rotation. [Col 2] The unmodified LEAP Hand (seen during training). [Col 3-7] Zero-shot embodiments with unseen graph variations and/or link length extensions (orange). We report (across 3 minute trial): 1) average cube angular velocity in degrees/second 2) real average velocity as a \% of sim average velocity for the same embodiment and policy (sim-to-real gap) and 3) number of times the cube fell off the hand.}
    \vspace{-2.5mm}
    \resizebox{1.8\columnwidth}{!}{%
    \begin{tabular}{c|c|c|c|c|c|c}
    \toprule
     \includegraphics[width=0.10\linewidth]{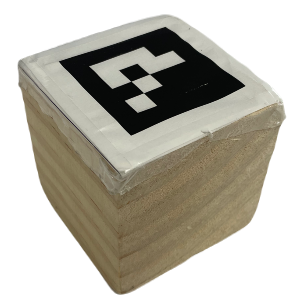}
     & \includegraphics[width=0.16\linewidth]{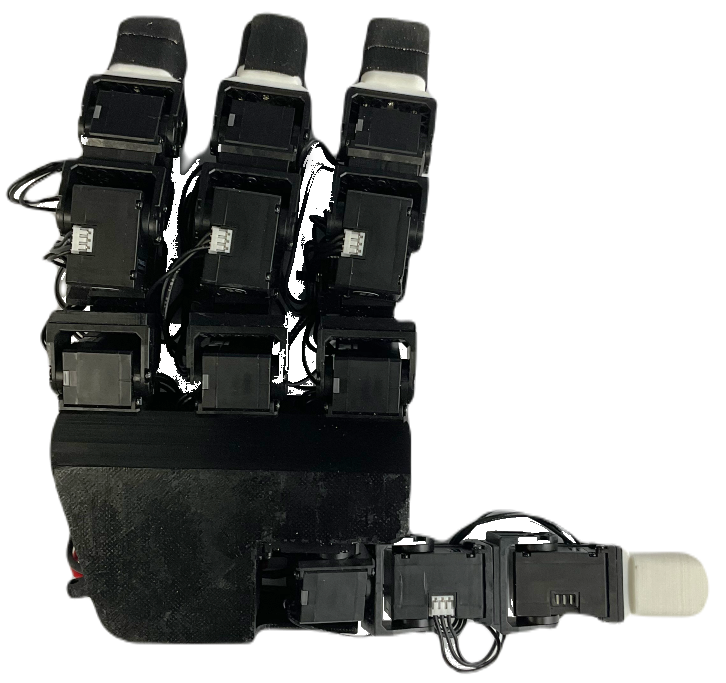}
     & \includegraphics[width=0.16\linewidth]{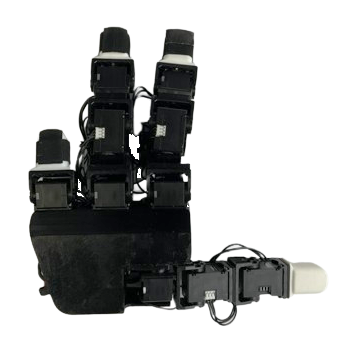}
     & \includegraphics[width=0.16\linewidth]{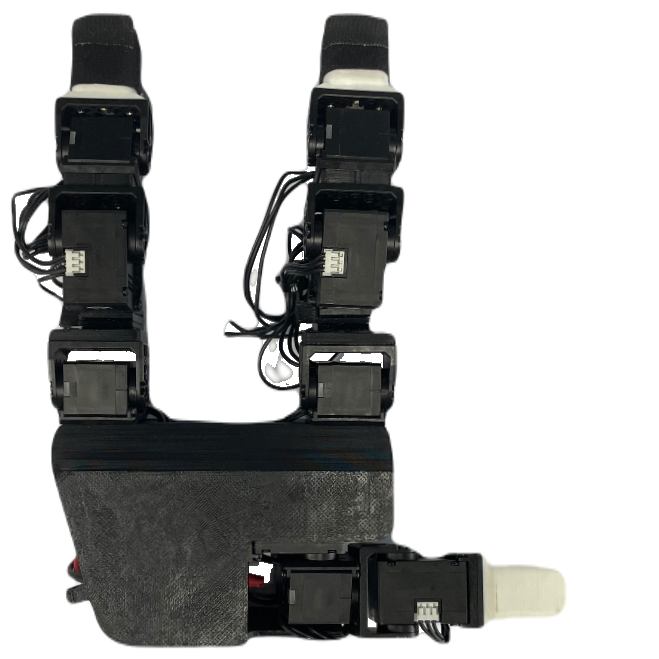}
     & \includegraphics[width=0.16\linewidth]{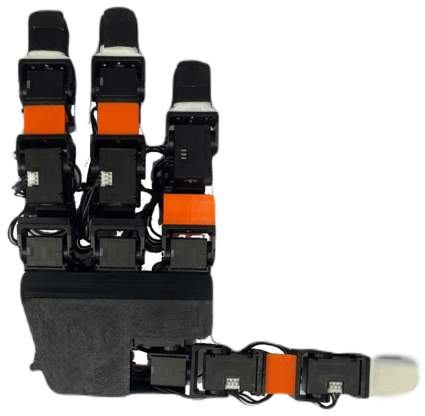}
     & \includegraphics[width=0.16\linewidth]{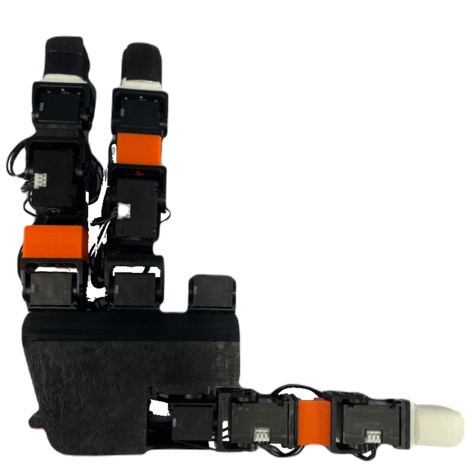}
     & \includegraphics[width=0.12\linewidth]{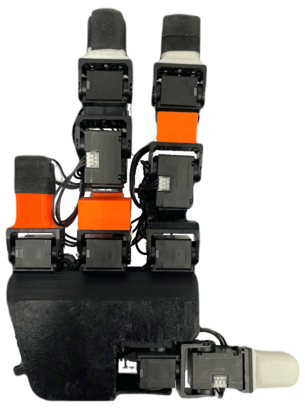}\\
     Variation & Training & New Graph & New Graph & New Geo & New Graph+Geo & New Graph+Geo \\
     ET    & 11.5, 55\%, 0  & 5.9, 50\%, 0 &  9.2, 93\%, 1 & 8.6, 57\%, 2   & 9.6, 67\%, 18 & -0.6, -6\%, 0 \\
     \ours & 20.5, 70\%, 0  & 8.3, 65\%, 0 &  21.8, 133\%, 0 & 19.0, 83\%, 1  & 9.0, 53\%, 16 & 1.2, 9\%, 0 \\
     \bottomrule
    \end{tabular}
    }
    
    \label{tab:realworld}
    \vspace{-4mm}
\end{table*}

\subsection{Key Results and Findings}

Table \ref{tab:simulation} shows \ours's performance on training embodiments (Training Graph) and on multiple types of zero-shot embodiments: unseen graph variations (New Graph), unseen link length variations (New Geo) and both (New Graph \& Geo). Link length extensions were \textit{never} seen during training.

\textbf{\ours~matches the performance of embodiment-specific experts on training embodiments.} Despite controlling 44 embodiments with graph changes using a \textit{single} set of network weights, \ours~reaches 99\% of the performance of the embodiment-specific RL experts (Tab.~\ref{tab:simulation}).

\textbf{\ours~improves zero-shot capabilities to graph and geometry variations.} Compared to the best performing baseline for each category, \ours~achieves a 16\% improvement with zero-shot to new graph, an 10\% improvement with zero-shot to link length variations, and a 20\% improvement with zero-shot to both graph and link length variations (Tab.~\ref{tab:simulation}).

\textbf{Self-modeling improves performance only with graph encoding present.} We observe that adding self-modeling (SL) to the baseline (ET) consistently decreases zero-shot performance by an average of 4.6\% across the three categories, but yields an average increase of 5.1\% when added to the model with graph encoding (ET+PE+SE) (Table~\ref{tab:simulation}). This result intuitively makes sense as the ET baseline has no graph encoding which is needed to complete the forward kinematics task and thus self-modeling provides a poor training signal. In the presence of graph encoding, forward kinematics serves as a useful loss to improve cross-embodiment transfer.

\textbf{Depth-first graph linearization lowers performance under graph variations.} The DFS linearization (ET+DFS) serves as a simple, yet incomplete graph representation as the DFS ordering is not unique. Additionally, if a joint earlier in the linearization is removed, all subsequent joints will shift down and receive a new positional encoding. When only link length variations are present, linearization improves zero-shot performance over no positional encoding (ET) by 11\% (Tab.~\ref{tab:simulation}). However, with unseen graph variations, performance drops by 26\%. This result intuitively makes sense as the linearized positional encoding overfit to training embodiment graphs, but fail to generalize to unseen graphs due to the aforementioned issue with DFS. This motivates the need for the more complete graph representation used in \ours.

\textbf{Spatial and parent-child encoding help policy learning.} Relative to ET with self-modeling (ET+SL), adding the parent-child graph encoding (ET+PE+SL) improves performance by 19\% and adding the spatial graph encoding (ET+SE+SL) improves performance by 25\% on average across zero-shot tasks. Adding the parent-child bias when the spatial bias is already present provides no statistically significant improvement, indicating that our directed encoding is not critical for performance, perhaps as parent-child relationships can be inferred from correlations in joint states.

\begin{figure}[t]
    \vspace{-1mm}
    \centering
    \includegraphics[width=0.6\linewidth]{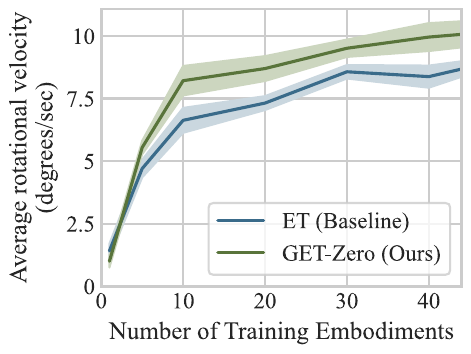}
    \vspace{-4mm}
    \caption{\textbf{Impact of training embodiments on zero-shot graph generalization}. We observe that even with fewer training embodiments, \ours~achieves reasonable in-hand rotation performance (5 seeds).}
    \label{fig:training_embodiment}
    \vspace{-7mm}
\end{figure}

\textbf{Zero-shot with fewer training embodiments.} To validate that zero-shot performance isn't solely due to a large number of training embodiments that might be similar to the zero-shot embodiments, we evaluate our method on smaller subsets of training embodiments as shown in Fig.{~\ref{fig:training_embodiment}}. \ours~achieves reasonable rotation performance even with only 10 training embodiments, indicating that zero-shot performance works when demonstration data is only available for fewer embodiments. The 44 training embodiments for results in Tab.~\ref{tab:simulation} is 19\% of the 236 total graph variation embodiments generated, though not all 236 embodiments are equally capable of completing the rotation task.

\textbf{Graph encoding improves forward kinematics prediction.} We train a GET model with only the forward kinematics head (no action head) to validate self-modeling capabilities for unseen embodiments. In Tab.~\ref{tab:fk} we observe that \ours~achieves much higher precision predicting per-joint positions due to the graph encoding mechanism.

\subsection{Real-world Evaluation}
We present results from a real-world evaluation of \ours~on unseen embodiments in Tab.~\ref{tab:realworld} following the same evaluation methods as for simulation. We observe GET-Zero outperforms the ET baseline for both the original LEAP Hand~\cite{leaphand} as well as unseen graph and link length variations. Empirically, the ET policy struggles to continuously rotate the cube, and we observe high finger precision is required for stable control. We observe a sim-to-real gap, but \ours~achieves zero-shot performance above simulation for the third embodiment. For another embodiment (second from right), a missing index finger causes the hand to drop the cube many times. For the right-most embodiment, a shortened index finger and thumb make it challenging for the hand to start the rotation cycle.

\begin{table}[t]
    \vspace{-1mm}
    \centering
    \caption{Average per-joint forward kinematics error in millimeters\\
    \vspace{-1mm}
    across 10 zero-shot graph variations (5 seeds)}
    \vspace{-2.5mm}
    \begin{tabular}{l|cccc}
    \toprule
                              & New Graph\\
    \toprule
     ET~\cite{mybodyisacage}  & 7.97$\pm1.36$\\
     ET+DFS~\cite{metamorph}    & 9.03$\pm$3.22\\
     \ours~(Ours)                    & \textbf{0.82}$\bm{\pm}$\textbf{0.81}\\
     \bottomrule
    \end{tabular}
    \vspace{-6mm}
    
    \label{tab:fk}
\end{table}

\subsection{Limitations}
\label{sec:limitations}

It's unlikely that \ours~trained on LEAP hands would zero-shot transfer to a new robot hand model. For example, we train with at most 16 joints and do not encode joint limit ranges, motor strength, friction properties or finger shape, all of which vary with a new hand and are important in manipulation tasks. Future work can extend \ours~to learn to iteratively improve a robot's design or explore if \ours~supports multi-task settings such as locomotion and manipulation with a single network.

\section{Conclusion}
\label{sec:conclusion}

We present \ours, an embodiment-aware model architecture and training procedure that enables zero-shot control of new robot designs. Through a case study on an in-hand object rotation task, we demonstrate the ability of our model to control a wide range of hardware configurations of a multi-fingered hand under variations in embodiment graph and geometry. Our results in simulation and real show that the graph encoding and self-modeling features in \ours~improve cross-embodiment transfer. We hope that \ours~serves as a useful, method for the robotics community to share knowledge across similar robot designs.

\section*{ACKNOWLEDGMENT}

We thank Kenneth Shaw, Prof. Pathak's lab at CMU, and the Prof. Liu's Movement lab at Stanford for sharing LEAP Hand hardware. We also thank Huy Ha, Xiaomeng Xu, Mengda Xu and Haochen Shi for their helpful feedback and fruitful discussions. This work was supported in part by Sloan Fellowship and NSF \#2132519.  The views and conclusions contained herein are those of the authors and should not be interpreted as necessarily representing the official policies, either expressed or implied, of the sponsors.

\bibliography{IEEEabrv,references}

\end{document}